\documentclass[letterpaper]{article}
\usepackage{aaai}
\usepackage{times}
\usepackage{helvet}
\usepackage{courier}
\usepackage{graphicx}  
\usepackage{lipsum}    
\usepackage{booktabs, multirow} 
\usepackage{soul}
\usepackage[table]{xcolor} 
\usepackage{changepage,threeparttable}

\usepackage{amsmath} 
\usepackage{braket}  
\usepackage{amsfonts} 

\usepackage[utf8]{inputenc}

\begin{document}
%
\title{Quantum Machine Learning in Climate Change and Sustainability: a Review
}
\author{Amal Nammouchi\textsuperscript{\rm 1}, Andreas Kassler\textsuperscript{\rm 2,3}, Andreas Theorachis\textsuperscript{\rm 4}\\
Karlstad University, Computer Science Department, 65635 Karlstad, Sweden\textsuperscript{\rm 1,2}\\
Karlstad University, Electrical Engineering Department, \\65635 Karlstad, Sweden\textsuperscript{\rm 4}\\
Deggendorf Institute of Technology,  Department of Applied Computer Science, \\94469 Deggendorf, Germany\textsuperscript{\rm 3}\\
amal.nammouchi@kau.se\textsuperscript{\rm 1}, andreas.kassler@kau.se\textsuperscript{\rm 2}, andreas.kassler@th-deg.de\textsuperscript{\rm 3}, andreas.theocharis@kau.se\textsuperscript{\rm 4}\\
\\
}

\maketitle
\begin{abstract}
\begin{quote}
Climate change and its impact on global sustainability are critical challenges, demanding innovative solutions that combine cutting-edge technologies and scientific insights. Quantum machine learning (QML) has emerged as a promising paradigm that harnesses the power of quantum computing to address complex problems in various domains including climate change and sustainability. 
In this work, we survey existing literature that applies quantum machine learning to solve climate change and sustainability-related problems. We review promising QML methodologies that have the potential to  accelerate decarbonization including energy systems, climate data forecasting, climate monitoring, and hazardous events predictions. We discuss the challenges and current limitations of quantum machine learning approaches and provide an overview of potential opportunities and future work to leverage QML-based methods in the important area of climate change research. 
\end{quote}
\end{abstract}

\noindent 



\section{Introduction and Background}
Climate change and global sustainability present pressing challenges, necessitating innovative solutions for managing complex distributed systems such as energy systems. While classical machine learning techniques have been applied to several problems in this area, Quantum machine learning (QML) offers a promising approach to overcome classical machine learning (ML) limitations in climate change research by leveraging quantum computing \cite{singh2021quantum}. 
This section introduces the need for significant actions to face climate change, most importantly, by introducing new cutting-edge technologies such as quantum machine learning (QML) \cite{book} to help accelerate the CO2-free transition. We present a brief overview of quantum machine learning fundamentals introducing quantum computing concepts and quantum neural network paradigms. The paper highlights the motivation and advantages of using QML to address challenges related to  mitigation and adaptation of climate change applications.
\subsection{Relation between QML and Climate Change}
The urgency to address climate change-related issues has reached a critical juncture, demanding immediate and innovative actions. With the planet experiencing unprecedented shifts in weather patterns, historically recorded highest temperatures and heat waves, rising sea levels, and ecological disruptions, the imperative to combat climate change has never been more evident. To effectively navigate this global challenge and expedite the transition to a sustainable future, harnessing cutting-edge technologies such as QML is an important step. Quantum machine learning presents a significant opportunity to better understand complex climate dynamics. Because QML can process and analyze intricate data sets at an unparalleled speed, better insights into climate models would be a significant advantage in enhancing predictive accuracy which allows  for more informed decision-making. As the climate crisis accelerates, integrating quantum machine learning into our efforts not only underscores our commitment to innovative problem-solving but also offers a powerful tool to drive the rapid changes required for a resilient and sustainable world.

\subsection{A brief Overview of Quantum Machine Learning Fundamentals}
\subsubsection{Quantum Computing}
is based on the principles of quantum mechanics, while classical computation is built on the rules of classical physics \cite{nielsen_chuang_2010}\cite{desurvire_2009}. Classical computers operate by manipulating bits, while in quantum computers, the information is processed via the means of its building blocks called qubits. Quantum bits or qubits live in a two-dimensional linear vector or Hilbert space, unlike bits that can assume discrete values of either 0 or 1.  The two computational basis states that span the Hilbert space of a qubit are denoted by the states $|0\rangle$ and $|1\rangle$ , as shown in Eq. (1). 
\begin{equation}
|0\rangle = 
\begin{bmatrix}
1 \\
0
\end{bmatrix}
, \quad
|1\rangle = 
\begin{bmatrix}
0 \\
1
\end{bmatrix}
\end{equation}

Quantum computing works on the basis of two principles of Quantum mechanics: superposition and entanglement. Superposition states that the quantum states can be added together and the qubits can be broken down into multiple quantum states. This principle allows the bit to be both one and zero or neither at any given time which simply represents a linear combination of its states:
\begin{equation}
|\psi\rangle = \alpha|0\rangle + \beta|1\rangle.
\end{equation}
Where coefficients $\alpha$ and $\beta$ are complex numbers ($\alpha, \beta \in \mathbb{C}$) and are often referred to as probability amplitudes such that:
\begin{equation}
| \alpha |^2 + | \beta |^2 = 1
\end{equation}
Where \( |\alpha|^2 \) is the probability of the qubit collapsing to the state \( |0\rangle \) upon measurement, and \( |\beta|^2 \) is the probability of it collapsing to the state \( |1\rangle \).
In theory, by leveraging the principle of superposition, a qubit can store and process information more rapidly than classical computers, offering potential energy efficiency advantages.



In addition to superposition, qubits also exhibit quantum entanglement allowing them to form co-relations between individually random behaviors. This property plays an important role in applications such as malicious attack detection, secure communication, and information processing. Interestingly, the qubits can be used to train machine learning algorithms in a much faster way as we will see later.

\begin{figure}[t!]
    \centering
    \includegraphics[width=\linewidth]{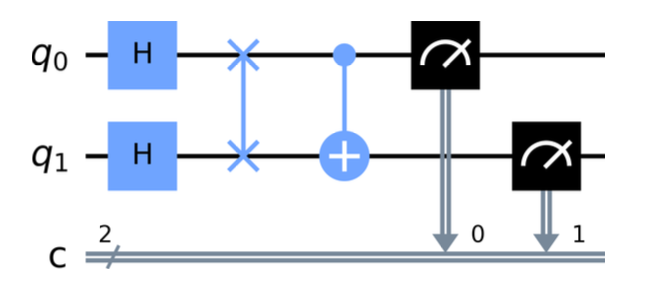} 
    \caption{Example of q quantum circuit with two qubits (generated using Qiskit
    )}
    \label{fig:quantum_circuit}
\end{figure}

 Classical computers manipulate bits using logic gates such as AND, OR, NOT, NAND, XOR, etc. Likewise, quantum computers manipulate qubits using quantum gates. A quantum gate is modeled as a unitary matrix that multiplies the system qubit state, and all quantum operations must be reversible (except measurement). A quantum program, implemented as a quantum circuit, is a sequence of quantum gates applied over one or more qubits as show in Figure \ref{fig:quantum_circuit}. More details about QC fundamentals can be found in \cite{qc}.

\begin{figure}[t!]
    \centering
    \includegraphics[width=\linewidth]{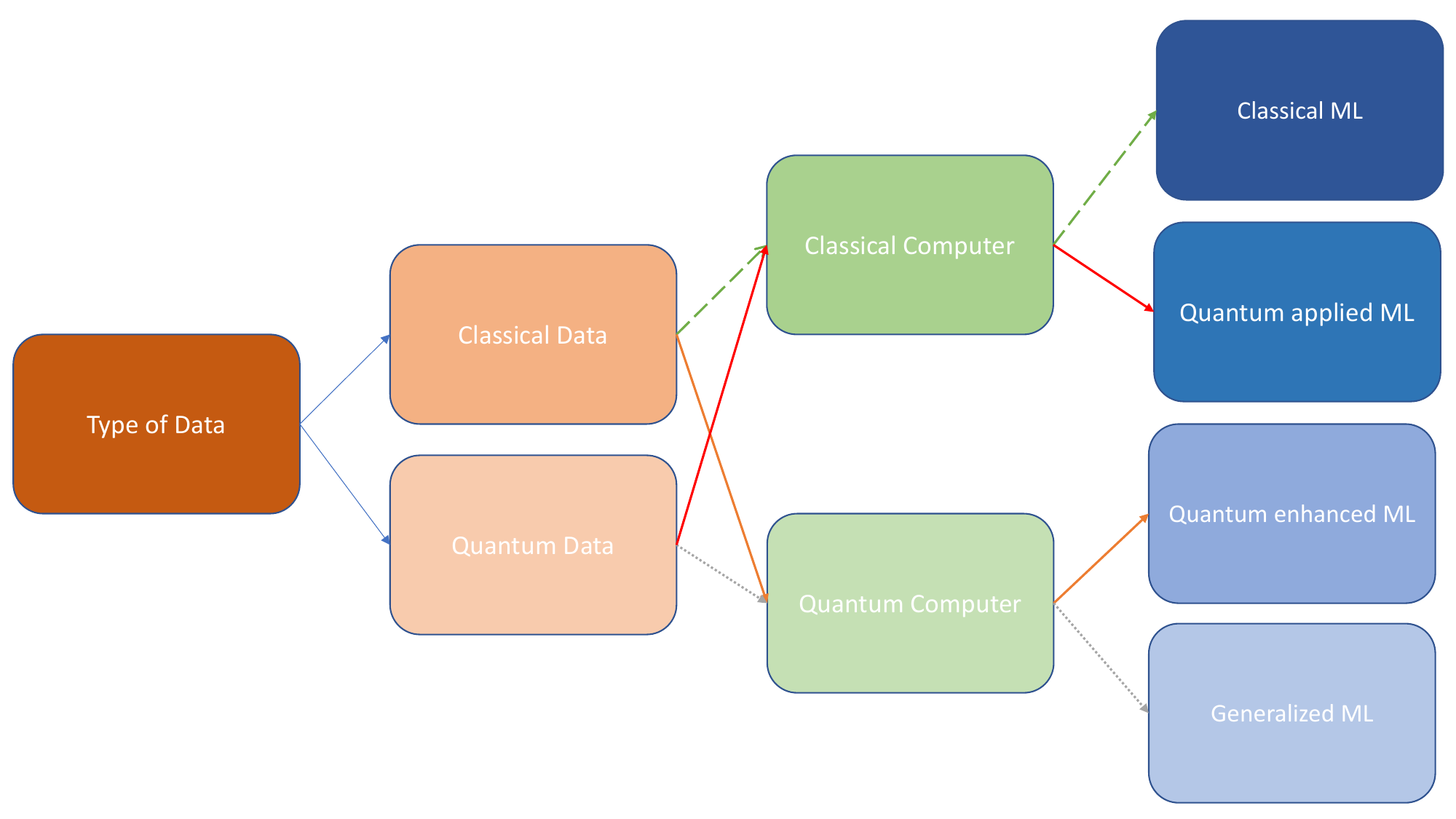} 
    \caption{Quantum Machine Learning Paradigms}
    \label{fig:QML}
\end{figure}
\subsubsection{Quantum Neural Networks}
Quantum machine learning is an emerging research area that bridges quantum computing and machine learning. 
Being at the border of these two research disciplines, QML entails methods that allow the exploitation of quantum phenomena to improve machine learning algorithms and the application of machine learning algorithms for improving quantum algorithms and designs. QML is based upon two main components — data and algorithms. They can be either quantum or classical as shown in Figure \ref{fig:QML} . 

\begin{figure*}[t!]
    \centering
    \includegraphics[width=0.7\textwidth]{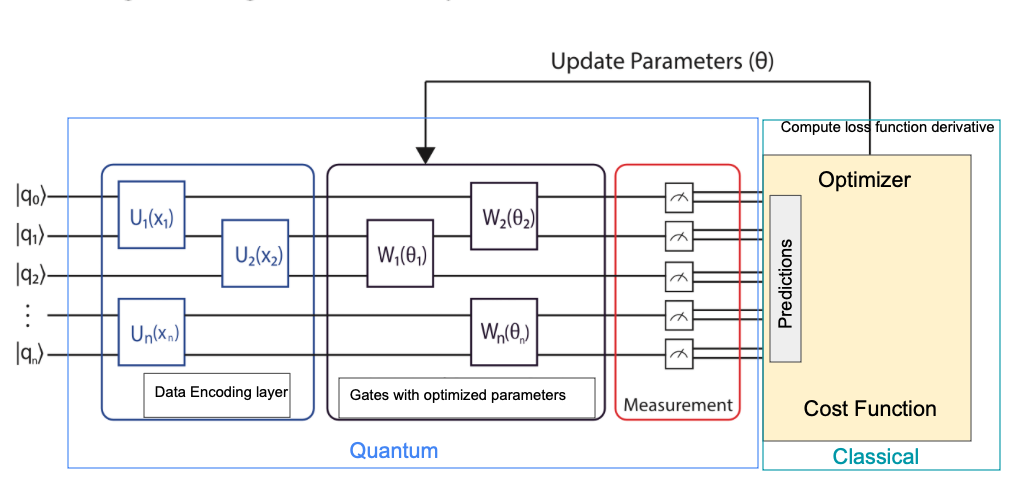} 
    \caption{General structure of hybrid quantum-classical QNN}
    \label{fig:QNN}
\end{figure*}

Quantum neural networks (QNNs) are currently one of the most trending topics in quantum machine learning. They represent a specific class of hybrid quantum-classical models that are executed in both quantum processors as well as classical processors to perform a single task. The QNN architecture has a structure that loosely resembles that of classical neural networks, as show in Figure \ref{fig:QNN}.
\subsection{The role of Quantum ML in Climate Change}
Classical ML has already played a crucial role in analyzing climate data and making predictions. As the complexity of climate models and the need for real-time decision-making grows, QML offers a new opportunity \cite{havenstein2018comparisons}. There are two primary issues that limit the performance of classical ML algorithms. The first pertains to the availability of high-quality training data, while the second revolves around the computational resources required to handle the immense volumes of data, which is common for climate models on a planetary scale. QML harnesses the unique properties of quantum computing to tackle complex problems more efficiently than classical computers. When applied to climate science, QML can enhance our understanding of climate patterns, improve climate modeling accuracy, and optimize strategies for mitigating climate change impacts. By leveraging quantum algorithms and quantum annealers, QML can process massive datasets, simulate intricate climate models, and optimize resource allocation for sustainable energy production. 


\section{Applications of QML for Climate Change}
In this section, we review existing literature that investigate the application of QML for climate change and sustainability related applications. The applications are grouped into 1) mitigation applications for decarbonization acceleration including energy systems, transportation and agriculture, 2) climate data forecasting applications including weather, load, renewable energy and carbon price forecasting, 3) climate monitoring applications including earth monitoring and satellite imagery and 4) climate applications for hazards prediction. A summary of the reviewed work is presented in Table\ref{tab:summary }.
\subsection{QML for Decarbonization Acceleration}
Decarbonization acceleration refers to efforts and applications aimed at rapidly reducing carbon emissions and transitioning to a low-carbon or carbon-neutral economy. In the context of climate change, decarbonization is a critical strategy to mitigate the impacts of global warming. Energy systems, transportation and agriculture form core applications for this transition. In this section we review literature that address this issue using QML techniques.

\subsubsection{Energy Systems} Transitioning from fossil fuels to renewable energy-based systems and establishing optimised sustainable management systems are pivotal steps towards achieving climate neutrality.
\cite{AJAGEKAR2022112493} reviewed the prospects of QC utilisation in general in various areas of applications in energy sustainability  to help address climate change. The authors present in depth review about different QC based algorithms that can be used for energy sustainability, including QML methods. Our work in this section differs as we review the literature that implemented QML for energy systems.\\
In a recent study by \cite{9670708}, a data-driven approach using multi-agent quantum deep reinforcement learning is proposed for distributed frequency control in islanded microgrids. It combines a traditional deep reinforcement learning (DRL) framework with QML for learning  the optimal cooperative control strategy. The proposed method reduces parameter requirements and improves training performance. These model improvements inherently enhance the efficiency of islanded microgrids, driving them closer to offering residents more eco-friendly services.
In \cite{en15166034}, the authors study the benefits and limitations of quantum reinforcement learning (QRL) to solve energy-efficiency scenarios in the contexts of  Heating, Ventilation, and Air Conditioning (HVAC) control, electric vehicle energy management, and profit optimization for charging stations. The results indicate that quantum neural networks offer improved accuracy and cumulative rewards compared to their classical counterparts while requiring fewer parameters for learning. However, the learning process is slower for QRL based models. This suggests the potential for QRL to enhance energy-efficient solutions in various domains.
In \cite{PV}, the authors present the implementation of a quantum neural network as a controller in a photovoltaic solar system.
The primary goal was to effectively track the maximum power point and subsequently transfer the panel's generated power to the load with minimal losses. The quantum-driven controller showcased rapid adaptability to a range of environmental conditions, outperforming classical methods thus ensuring better optimization in solar energy systems. 
\cite{LI2023121181} introduced a distributed area autonomy load frequency control (DAA-LFC) technique, designed for the efficient balance of interests among multiple grid operators while ensuring swift frequency restoration in multi-area microgrids.  The authors propose a quantum-based algorithm, termed DQMA-DMDPG, which combines large-scale and meta-learning for multitask collaboration by adjusting exploration parameters for varied tasks. The effectiveness of this method in minimizing frequency deviations, slashing power generation expenses, and aligning the interests of different operators was empirically validated. In the context of smart generation control, \cite{YIN2022109804} proposed lightweight robust quantum Q-learning (LRQQL) methods to solve the problems of weak self-adaptation and low convergence rate of current SGC methods of zero-carbon power systems. The authors verify the feasibility of LRQQL algorithms which proves to outperform classical methods such as state-action-reward-state-action (SARSA) and Q-learning in terms of control error, frequency error and convergence rate.

\subsubsection{Transportation}
Transportation and vehicle electrification has recently been significantly influenced  by the economic and environmental issues related with fossil fuel transportation worldwide \cite{ev}. Recent advancements in QML are beginning to  influence research directions in this field.
\cite{GEETHA2023108273} proposes a hybrid approach combining the Eurasian Oystercatcher Optimizer (EOO) and Quantum Neural Network (QNN) for jointly optimizing the placement of electric vehicle charging stations (EVCS) and capacitors within distribution systems (DS). The proposed EOO-QNN method aims to regulate capacitors to maintain voltage profiles, increase net gain, and reduce active power loss. The authors report that the proposed technique is more stable for increasing the voltage compared with commonly used optimization methods including Salp Swarm Algorithm (SSA), and Particle Swarm Optimization (PSO). In \cite{10039707}, the authors introduced BQL-ET, which is a Blockchain and QRL-based model  for optimized energy trading in the context of e-mobility using microgrids (MGs). It addresses the challenge of setting the selling price for electricity generated by microgrids that charge Electric Vehicles (EVs) by proposing a double-auction mechanism to determine optimal market-trading prices. The model employs smart contracts on a consortium blockchain to evaluate overall utility, transforming the problem into a Markov Decision Process (MDP) and using (QRL) for policy development. The results demonstrate that BQL-ET converges faster, maximizes utility for both microgrids and EVs, and achieves lower transaction confirmation times and optimal market-trading prices compared to existing models. This result in using MGs to charge (EVs) in a more optimal way, which is important for further pushing electric mobility forward.
\begin{table*}[t!]\centering
\caption{Summary of the literature review}\label{tab:summary }
\scriptsize
\begin{tabular}{p{2.8cm} p{2.3cm} p{2.3cm} p{2.3cm} p{5.5cm} p{0cm}}\toprule
Application Category &Climate Application &Reference &QML approach &Description \\\cmidrule{1-5}
\multirow{5}{*}{Decarbonization Acceleration} &\multirow{3}{*}{Energy Systems} & \cite{9670708} & Multi-agent QDRL & Distributed
frequency control in islanded microgrid& \\
&  & \cite{en15166034}&QRL & Energy efficient optimisation in HVAC control, EV and energy management \\
& &\cite{PV} &QNN &Photovoltaic solar system controller to track maximum power point . \\
& &\cite{LI2023121181} &DQMA-DMDPG &Distributed area
autonomy load frequency control\\
& &\cite{LI2023121181} &Quantum Q-learning & Smart grid energy and frequency control\\

&Transportation &\cite{GEETHA2023108273} & hybrid QNN & Optimise the
placement of electric vehicle charging stations (EVCS) and
capacitors within distribution systems \\
& &\cite{10039707}
 & hybrid QRL & Optimise energy trading between EVs and MGs.& \\

&Agriculture &\cite{agriculture} & QCNN&Prediction of wheat plant disease &  \\\cmidrule{1-6}
\multirow{8}{*}{Data Forecasting} &\multirow{3}{*}{Weather Forecasting} &\cite{weathersvm} &QSVM & Solar irradiation forecasting& \\
& & & & & \\
& & & & & \\
&\multirow{3}{*}{RE Forecasting} &\cite{10109785} &hybrid LSTM &One day-ahead spatio-temporal wind speed forecasting  & \\
& &\cite{10111033} &QLSTM & Solar irradiation forecasting for solar energy prediction& \\
& & & & & \\
&Carbon Price Forecasting & \cite{Cao2023}&L-QLSTM &Carbon Price forecasting & \\
&Load Forecasting & \cite{elmen}&Q-Elman &Load forecasting & \\\cmidrule{1-6}

Climate Monitoring &- &\cite{AYOADE2023337} &QML classifier &Satellite-observed hyperspectral images classification, distinguishing vegetation from other land type & \\
&- &\cite{sebastianelli2021circuitbased}  &QCNNs & Image classification for remote sensing application& \\
&- &\cite{gupta2023potential} 
 &POK &Assess the impact of PQK features on multispectral classification accuracy using satellite imagery & \\
&- & \cite{9323065}

 &VQC &Satellite imagery classification & \\
\cmidrule{1-6}

Hazards Prediction &- &\cite{9848250} &QSVM &Earthquake prediction & \\
&- & \cite{10188662}&VQC &Asteroids hazards prediction & \\
&- &\cite{10198241} & & & \\
\midrule
\bottomrule
\end{tabular}
\end{table*}


\subsubsection{Agriculture}
Modern advancements in deep learning, both classical and quantum, have significant potential when applied to the field of agriculture. A study by \cite{agriculture} offers a detailed comparison of these methods, particularly in the prediction of wheat plant diseases. Their primary aim is to empower stakeholders with the insights needed for informed decision-making in future agricultural strategies. In their research, four distinct neural network models were evaluated: Convolutional Neural Network (CNN), Neural Network (NN), Quantum Neural Network (QNN), and Quantum Convolutional Neural Network (QCNN). The classical CNN model outperformed the quantum models achieving an accuracy of 91.32\%. This outcome challenges the prevailing belief in the superiority of quantum-based approaches for ML tasks. This suggests that without the infrastructure of a genuine quantum computing platform, the potential advantages of QNN might remain unrealized on conventional hardware. 
These findings emphasize a crucial observation: while quantum methods are undeniably promising, they don't necessarily and always outperform classical methods, especially when deployed on non-quantum systems.  In fact, quantum algorithms are designed to harness the unique properties of quantum mechanics, such as superposition and entanglement, which don't have direct counterparts in classical systems. Thus, when quantum methods are implemented or simulated on classical systems (quantum simulators), they can be inefficient and don't exploit the potential advantages of genuine quantum computation. 
\subsection{QML for data forecasting}
Data forecasting of time-series data in the context of climate change has many important use cases.  Data forecasting can be used to predict the weather, the power load, renewable energy production, and carbon prices. The accuracy of such forecasts is imperative for high-quality decision-making, especially  in smart MGs. Accurate and timely forecasting helps policymakers, businesses, and the public to prepare for and mitigate the impacts of climate change. Quantum computers, as recent studies suggest, might be adept at perceiving trends in time series data more effectively than classical methods \cite{ccc}\cite{kaushik2022onestep}. In this section, we review the related studies. 

\subsubsection{Weather Forecasting}
\cite{can} discuss the fundamental limitations that impede the  performance increase of supercomputers based on silicon transistors. The limitations in computing power severely impact the development of accurate numerical weather and climate prediction models. The authors showcase how different global meteorological centers are pushing the boundaries of forecast accuracy and model resolution at the expense of excessive computing power required for such models. 
\cite{weathersvm} propose a quantum support vector machine (QSVM) algorithm for forecasting solar irradiation which is useful for renewable energy production prediction and weather forecasting. 
\cite{supp} explored the impact of quantum phenomena, specifically superposition and entanglement, on weather forecasting using a variational quantum circuit (VQC) model. Incorporating the entanglement layer between the variational layers has made significant improvements in the circuit performance in this study. Additionally, the use of the superposition layer before the data encoding layer resulted in the use of less variational layers which improves the performance. This study highlights the potential synergy of Quantum Neural Networks (QNN) when combined with other techniques, leading to highly accurate models that can calculate weather forecasts very fast.


\subsubsection{Renewable Energy Forecasting}
\cite{10109785} introduced a novel hybrid  model combining quantum processes and residual Long Short-Term Memory (LSTM) networks. The proposed model, optimized by particle swarm optimization (PSO), aims at one-day-ahead spatiotemporal wind speed forecasting. The proposed method outperforms traditional methods, such as GRU, KELM, CNN, BLSTM, SVRM, and ANN, in terms of accuracy. One limitation of the proposed method is the extended training duration which is attributed to the inclusion of the quantum embedding layer and the exclusive utilization of a quantum simulator. Nevertheless, the proposed method is fast enough for 24-hour ahead wind power and speed forecasting.\\
Due to the randomness of solar energy due to short varying weather phenomena, the output of the PV system will fluctuate. This form of randomness imposes a challenge on forecast accuracy, which will affect the safe operation of the grid. To improve forecast accuracy,  a high-precision hybrid prediction model based on variational quantum circuit (VQC) and long short-term memory (LSTM) network is developed by \cite{10111033} to predict solar irradiance one hour in advance. The authors propose a novel hybrid quantum long short-term memory (QLSTM) model architecture, which embeds the VQC into the LSTM. The proposed QLSTM model is compared with the commonly used SARIMA, CNN, RNN, GRU, and LSTM models under different experimental conditions. The results show that the overall performance of the QLSTM model is better than the baseline models in terms of annual average mean absolute error and similar metrics. 

\subsubsection{Load Forecasting} Accurate load forecasting is crucial for the optimisation and operational stability of smart grids and RE-based MGs. In this context, \cite{elmen} propose a short-term load forecasting model based on quantum Elman neural networks. The authors demonstrate that the accuracy is higher compared with classical models based on conventional Elman neural network and conventional feed-forward neural network. 

\subsubsection{Carbon Price Forecasting} Carbon pricing is a tool that puts a monetary value on carbon emissions, incentivizing businesses and individuals to reduce their carbon footprint. Accurate forecasting of carbon prices allows businesses to make informed decisions about investments in cleaner technologies and practices, knowing the potential costs of continued high emissions.
\cite{Cao2023} introduced a hybrid quantum computing framework, L-QLSTM, aimed at predicting carbon prices. This framework builds upon the QLSTM model by incorporating linear layers, which enhance the quantum model's ability to learn. L-QLSTM provides comparable accuracy to classical LSTM for carbon price predictions. Its improved performance comes from the integration of linear embedding layers and an optimized variational quantum circuit. The study serves as one of the pioneering efforts to predict carbon prices using hybrid quantum methods.
\subsection{QML for climate monitoring}
Satellite image classification provides invaluable data that can be used to monitor the planet's health, validate scientific models, plan responses, and guide policy decisions related to climate change. Given the need to help expand the processing techniques to deal with a large volume of high-resolution data, Earth Monitoring can benefit from new and innovative Big-Data computational technologies. In this context, Some researchers tried to investigate the potential of quantum computing and QML for earth and climate monitoring-related applications. 
\cite{AYOADE2023337} demonstrated the application of QML in classifying satellite-observed hyperspectral images, distinguishing vegetation from other land types. 
\cite{senokosov2023quantum} utilizes quantum effects through hybrid quantum-classical approaches to further enhance the capabilities of traditional classical models. They propose two hybrid quantum-classical models: a neural network with parallel quantum layers and a neural network with a quanvolutional layer, which effectively address image classification problems.
\cite{sebastianelli2021circuitbased} developed a hybrid Quantum Convolutional Neural Networks (QCNNs) model for image classification for remote sensing applications. The authors underline the potential of applying quantum computing to an Earth Observation case study and demonstrate that the QCNN performance is higher than the classical methods. Additionally, the authors study various quantum circuits to show that the ones exploiting quantum entanglement achieve the best classification scores.
In another comparative study, \cite{gupta2023potential} evaluated the performance of several classical machine learning algorithms on the stilted re-labeled dataset of the Copernicus Sentinel-2 mission, when the algorithm has access to Projected Quantum Kernels (PQK) features. POK are a family of kernels that work by projecting the quantum states to an approximate classical representation.
The results from this study indicate a marked increase in classification accuracy with the integration of PQK features.
\cite{9323065} explored the application of quantum circuit-based neural network classifiers for multi-spectral data classification to gather land cover information. However, it is noteworthy that the overall classification score in this study lags behind what state-of-the-art classification systems can achieve. 
\subsection{QML for climate hazards predictions}
Earthquakes rank among the most devastating natural hazards, often leading to widespread destruction and significant loss of life. The importance of developing early warning systems for such natural calamities cannot be overstated. In a study by Dhotre et al. (2022), the potential of QSVM is explored for earthquake prediction. This effort underscores the continued search for efficient predictive tools in the realm of natural disaster mitigation. 
A separate study by \cite{10188662} presents a QML-based approach specifically tailored for predicting asteroid hazards. By utilizing Variational Quantum Circuits (VQC) and the Pegasos QSVC algorithm, the proposed method showed superior performance, registering an impressive accuracy of 98.11\% and an average F1-score of 92.69\%. Such advancements in QML-based asteroid hazard prediction can significantly enhance real-time risk detection and mitigation, playing a crucial role in safeguarding our planet's biodiversity.



\section{Discussion, Challenges, and Opportunities}

Quantum computing (QC) has recently demonstrated remarkable progress by achieving exceptionally high advantages in computational performance for certain tasks performed on quantum computers that are otherwise intractable to tackle even with the most powerful supercomputers.
Naturally, researchers aim to integrate QC and QML in climate change use cases and sustainability-related applications. Quantum-based optimization, quantum time series forecasting, and quantum image classification-based models have recently gained the most attention from researchers in the field. 
In the context of climate change, hybrid classical-quantum approaches or equivalent of classical ML models on a quantum simulator are most often used. Some of these models include  QDRL, QSVM, QNN and CQNN. These models typically 1) improve the accuracy and, 2) reduce the number of parameters  compared to classical ML-based methods. The impact of such improvements may help to accelerate the transition towards climate neutrality. However, these models were reported to be often slower. In fact, hybrid algorithms that combine both quantum and classical components can introduce overhead and additional operations. Additionally, the reviewed literature uses quantum simulators which are software tools that mimic the behavior of a quantum computer but run on classical computers. Simulators are useful when access to actual quantum hardware is limited, however, simulating larger quantum circuits can be resource-intensive and slow on classical hardware. More details about the limitations that lead to slower performance of hybrid or quantum models compared to their classical counterparts are discussed in the next subsection. Furthermore, for the study case of agriculture in \cite{agriculture}, the accuracy of the quantum model was reported to be worse than its classical counterpart. 

Several of the reviewed work was motivated by the energy-efficiency that quantum computing offers. This is in contrast to large ML models, which require an intensive amount of compute power and thus energy for their training leading to significant CO2 emissions required for producing those models. Quantum computers, could, in theory, solve certain problems more efficiently than classical computers, leading to energy savings. However, it should be noted that current quantum computers require extremely low temperatures to operate, which in itself requires significant energy for cooling. As technology advances, it is hoped that more energy-efficient quantum computers will emerge.

Common limitations of the studied literature are 1) restricted data encoding and quantum kernel usage, 2) computing on  ideal simulated quantum computer such as IBM-quantum\footnote{https://docs.quantum-computing.ibm.com} and PennyLane\footnote{https://pennylane.ai}, and 3) the assumption of an infinite number of samples sampled from the quantum circuits that encode the data. Especially, the last point implies that when researchers are working with QML algorithms, they might often assume (for the sake of simplicity or to overcome present-day technological constraints) that they can obtain an endless number of measurement samples from these quantum circuits that encode classical data.


\subsubsection{QML Challenges}

As QML is still in the early days, there remain several open issues in QML research. We summarise some of the major challenges below:

\begin{itemize}
\item \emph{ Linear-Non-linear Compatibility}: One of the main challenges for QNNs remains the linear-non-linear compatibility between neural network computation and quantum mechanics. Neural network computation is done in a non-linear fashion, that is, the activation function that triggers each neuron is non-linear, otherwise, the idea of layers in neural networks would serve no purpose. On the other hand, quantum systems behave in a linear way, which gives rise to the first incompatibility. 
\item \emph{Qubit Coherence}: The major challenge in quantum device development is that the qubits will lose quantum properties due to de-coherence. This means that qubits will become classical bits when these are in a superposition state. This problem may occur in mid-training which produces errors while trying to compute specific values. Naturally, environmental factors would cause noisy intermediate states that disturb the quantum devices. De-coherence can be addressed by protecting qubits from vibrations or by keeping them extremely cold. Many researchers are trying to find approaches to address this research challenge by developing error correction mechanisms. A practical trick is to keep the width of the circiuit small (i.e the number of layers).
\item \emph{Limited Qubits}: The number of qubits in quantum computers today is still limited. This makes it challenging to handle large-scale problems that require more qubits than are available.
\item \emph{Hybrid Approaches}: several proposed QML algorithms are hybrid, meaning they involve both quantum and classical machine learning steps. The back-and-forth can introduce latency, especially when considering cloud-based quantum processors.

\item \emph{Noise}: Current quantum devices are noisy and error-prone. Error correction or mitigation strategies are often needed, which can introduce overhead.

\item \emph{Circuit Depth and Width}:
Even if there are fewer parameters in a QML method, the quantum circuits required might be deep (many sequential gates) or wide (many qubits). Both depth and width can impact the feasibility of running a quantum circuit on today's quantum devices due to noise and qubit limitations.
\item \emph{Measurement Sampling}: Quantum systems provide probabilistic outcomes. To get reliable results, many measurements (samples) are often needed, introducing a time overhead.
\item \emph{Data Encoding Overhead}: Quantum computers operate on quantum data. Classical data must be encoded into quantum states, which can be a time-consuming process depending on the encoding scheme.

\end{itemize}

\subsubsection{Future Work and Potential Directions} It is important to note that QML is still in its early stages. Still, many steps are required  in order to reach the QC supremacy. More advanced models should be investigated in the context of climate change. Some of the potential applications are: 
\begin{itemize}

\item \emph{Climate Systems Modeling}: QML might significantly speed up climate simulations, enabling more accurate and granular forecasts.
\item \emph{Material Simulation}: QML can aid in simulating materials at the quantum level. This has implications for designing new materials that can absorb or reflect certain wavelengths of light, playing a role in climate control.
\item \emph{Climate Data Analysis}: With the increasing complexity and volume of climate data, QML can be used to identify patterns, anomalies, or trends that classical algorithms might miss or take too long to process.
\item \emph{Extreme Events}: Predicting extreme climate events like hurricanes, droughts, or floods with higher accuracy.
\item \emph{Weather Forecasting}: long-time horizon forecasting often suffer from low accuracy, and it is still an ongoing research even for the ML community. QMLs higher accuracy can be leveraged to enhance predictive models and provide more accurate, longer-term forecasts.
\end{itemize}

\section{Conclusion}



In this article, we surveyed various studies that employ QML-based methods to address climate change-related applications. Most of these studies demonstrate QML's efficiency, highlighting its advantages over classical ML methods in tackling specific climate-related challenges. It is crucial to emphasize that QML is still an emerging field. Continuous research aims to refine its capabilities in order to achieve quantum computational supremacy. As quantum computing (QC) hardware becomes more advanced and QML research progresses, the climate change research community should proactively harness these advancements, accelerating our journey toward climate neutrality.

\section{ Acknowledgments}
Parts of this work have been funded by the swedish energy agency (energimyndigheten) through the project AI4-ENERGI. Additional funding has been provided by the Bavarian Ministry of Science and Arts through the HighTech Agenda (HTA).

\bibliography{bibliography.bib} 
\bibliographystyle{aaai}



\end{document}